\documentclass{amsart}
\usepackage{amsthm}
\usepackage{amsmath}
\usepackage{amssymb}
 
\usepackage{tikz}
\usepackage{tikz-3dplot}
\usetikzlibrary{matrix,chains,positioning,decorations.pathreplacing,arrows, calc}
\usetikzlibrary{fit,backgrounds,intersections}
\usetikzlibrary{3d,decorations.text,shapes.arrows}
\usetikzlibrary{arrows.meta}
\usetikzlibrary{math}
\usetikzlibrary{fadings}
\usetikzlibrary{shapes.geometric}
\usetikzlibrary{bayesnet}
\usepackage{pgfplots}
\usepgfplotslibrary{patchplots}
\pgfplotsset{compat=1.9} 
\usepackage{graphicx}

\usepackage[
backend=biber,
style=numeric-comp
]{biblatex}
\usepackage[
bookmarksopen,
bookmarksdepth=2,
colorlinks=true,
linkcolor={blue},
citecolor={blue},
urlcolor=blue]{hyperref}

\DeclareGraphicsExtensions{.PNG} 
\begin{document}

\title{Expressive Power and Loss Surfaces of Deep Learning Models}
\author{Simant Dube}
\thanks{This article is an excerpt of and is based on a chapter from a recently published textbook on AI by the author, titled ``An Intuitive Exploration of Artificial Intelligence: Theory and Applications of Deep Learning"~\cite{SimantBook}. \\Email: simantdube@iitdalumni.com}
\address{{\rm Simant Dube is a computer scientist working in San Francisco area and is a former academic.}}
\email{simantdube@iitdalumni.com}

\begin{abstract}
The goals of this paper are two-fold. The first goal is to serve as an expository tutorial on the working of deep learning models which emphasizes geometrical intuition about the reasons for success of deep learning. The second goal is to complement the current results on the expressive power of deep learning models and their loss surfaces with novel insights and results. In particular, we describe how deep neural networks carve out manifolds especially when the multiplication neurons are introduced. Multiplication is used in dot products and the attention mechanism and it is employed in capsule networks and self-attention based transformers. We also describe how random polynomial, random matrix, spin glass and computational complexity perspectives on the loss surfaces are interconnected.
\end{abstract}

\maketitle

\section{Introduction}
In the last decade, there has been proliferation of applications of deep neural networks in the general field of Artificial Intelligence (AI) and specifically in computer vision, speech recognition and natural language understanding. Though there has been significant progress in practical techniques based on impressive empirical work, there has been expression of need to understand the theory behind the success of deep learning.

One of the comments on AI has been that we don't really understand what's going inside the black box of an AI model (in this article, an AI model refers to a deep learning model). What are all the hidden neurons doing when a CNN is recognizing a cat? How is an AI model able to generalize to unseen examples? Another persistent comment has been that we don't really know what's happening during the training process. What is the nature of this optimization landscape? Why doesn't the training get stuck in a local minimum? 

There are two goals of this article. The first goal is to serve as an expository tutorial in order to describe the following:
\begin{enumerate}
    \item How a deep neural network carves its target manifold and fits a function on it?
    \item Why SGD is effective in avoiding getting trapped in a local minimum?
\end{enumerate}
The second goal is to provide new results and insights which complement the current results. In particular, we show the following:
\begin{enumerate}
    \item An easy to understand geometrical mechanism illustrating how AI models carve out manifolds and fit functions.
    \item The generalization of the geometrical process when neurons can multiply. Multiplication occurs in the dot product operation and in attention which are used often in AI models. The AI model computes a piecewise continuous function over curved pieces whose number is exponential in the depth of the model. If it is a ReLU network in which the standard neurons can multiply, then the function is a piecewise polynomial function. In practice, the function will be more complex because of additional non-linearities which are added in the middle of networks such as capsule networks and transformers.
    \item The confluence and interconnection of several perspectives in understanding why SGD is able to find a good solution. We discuss random polynomial and computational complexity perspectives in addition to known random matrix and spin glass perspectives.
\end{enumerate}
The article is an excerpt of and is based on a recently published textbook by the author, see~\cite{SimantBook}.

\begin{figure}
\centering
\begin{tikzpicture}
    \node[anchor=south west] (image) at (0,0) 
        {\includegraphics[trim=70 150 0 150,clip,width=0.35\textwidth]{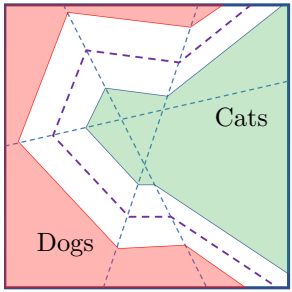}};
    \begin{scope}[x={(image.south east)},y={(image.north west)}]
        \node[] at (0.7,0.6) {Cats};
        \node[] at (0.2,0.2) {Dogs};
    \end{scope}
\end{tikzpicture}

\caption{AI chops off manifolds into convex polytopes. The dashed lines are the chiselling cuts.}
\label{fig:convex_polytopes}
\end{figure}

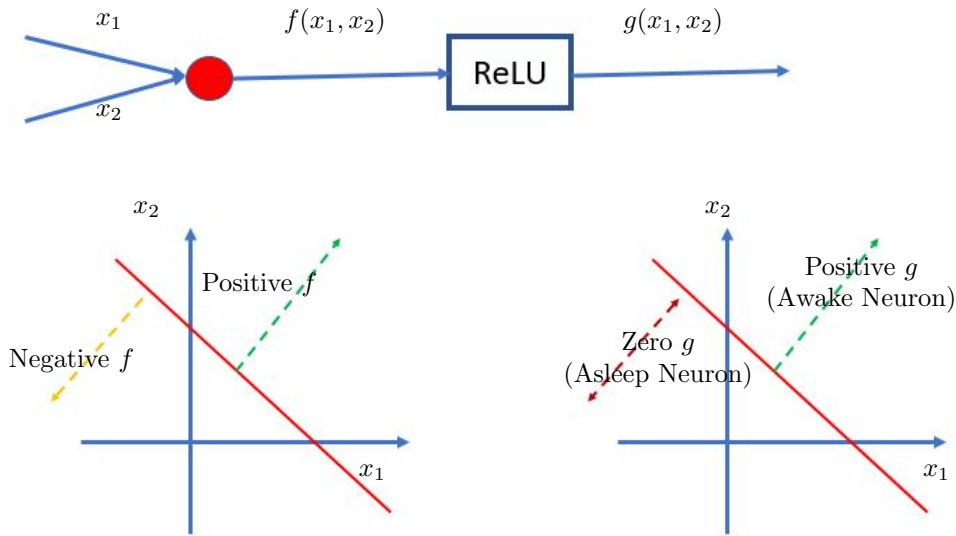
\begin{figure}[t]
\centering
\begin{tikzpicture}
\begin{scope}[draw=blue]

\node (manifold) at (3,0)
    {\includegraphics[width=\textwidth]{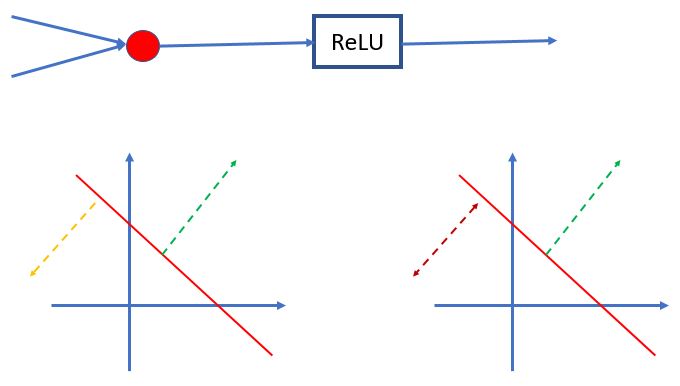}};

\node at (-2,3.5) {$x_1$};
\node at (-2,2.3) {$x_2$};
\node at (1,3.5) {$f(x_1, x_2)$};
\node at (5.5,3.5) {$g(x_1, x_2)$};

\node at (-1.5,1) {$x_2$};
\node at (1.5,-2.5) {$x_1$};
\node at (6.1,1) {$x_2$};
\node at (9,-2.5) {$x_1$};

\node at (0,0) {Positive $f$};
\node [align = center] at (8,0) {Positive $g$\\ (Awake Neuron)};

\node at (-2.5,-1) {Negative $f$};
\node [align = center]  at (5.3,-1) {Zero $g$ \\(Asleep Neuron)};


\end{scope}
\end{tikzpicture}

\caption{Activation region for a single ReLU neuron in its input space. The ReLU activation function zeros out the negative half-plane of $f$. We say that the neuron is awake or active in the positive half-plane and it is asleep or inactive in its other half-plane. Functions $f$ and $g$ assume 0 value at the red lines. In $d$-dimensional space, the lines are hyperplanes. Function $f$ is a linear function and $g$ is non-linear. Function $g$ is piecewise linear.}
\label{fig:ReLU_neuron}
\end{figure}

\section{Convex Polytopes}
As shown in Figure~\ref{fig:convex_polytopes}, the basic carved unit of AI is a convex polytope. We use the informal terminology of ``carving" and ``chiselling cuts" in analogy with a sculpting process. To understand how the input space gets divided into these pieces, we will start with a single ReLU neuron. See Figure~\ref{fig:ReLU_neuron} which shows how a ReLU neuron divides up the input space into two regions. Each region (half-space) is an infinite convex polytope. Each of the regions has an activation pattern associated with it. For a single neuron, it is either Active or Inactive.
Suppose the same input $(x_1, x_2)$ is sent to a second ReLU neuron. Now we will have four convex polytopes (unless the cutting hyperplanes are identical). Each convex polytope will have an activation pattern:
\[(a_1, a_2)\]
where $a_i \in \{\text{Active, Inactive}\}$ is the activation for neuron $i \in \{1,2\}$.

Here is another example. Consider a fully connected FFN with five input neurons, three hidden neurons and one output neuron, see Figure~\ref{fig:one_layer_ffn}. Each hidden neuron divides up the 5-D input space into two 5-D half-spaces. All three hidden neurons divide up the 5-D input space into many convex polytopes, each polytope having an activation pattern:
\[(a_1, a_2, a_3).\]
There is one output neuron. What does it do to the input space? Though it is clear that it will divide up the 3-D feature space $(h_1, h_2, h_3)$ into two 3-D half-spaces with a hyperplane, it is not obvious what its impact on the 5-D input space is.

\begin{figure}[ht]
\includegraphics[width=\linewidth]{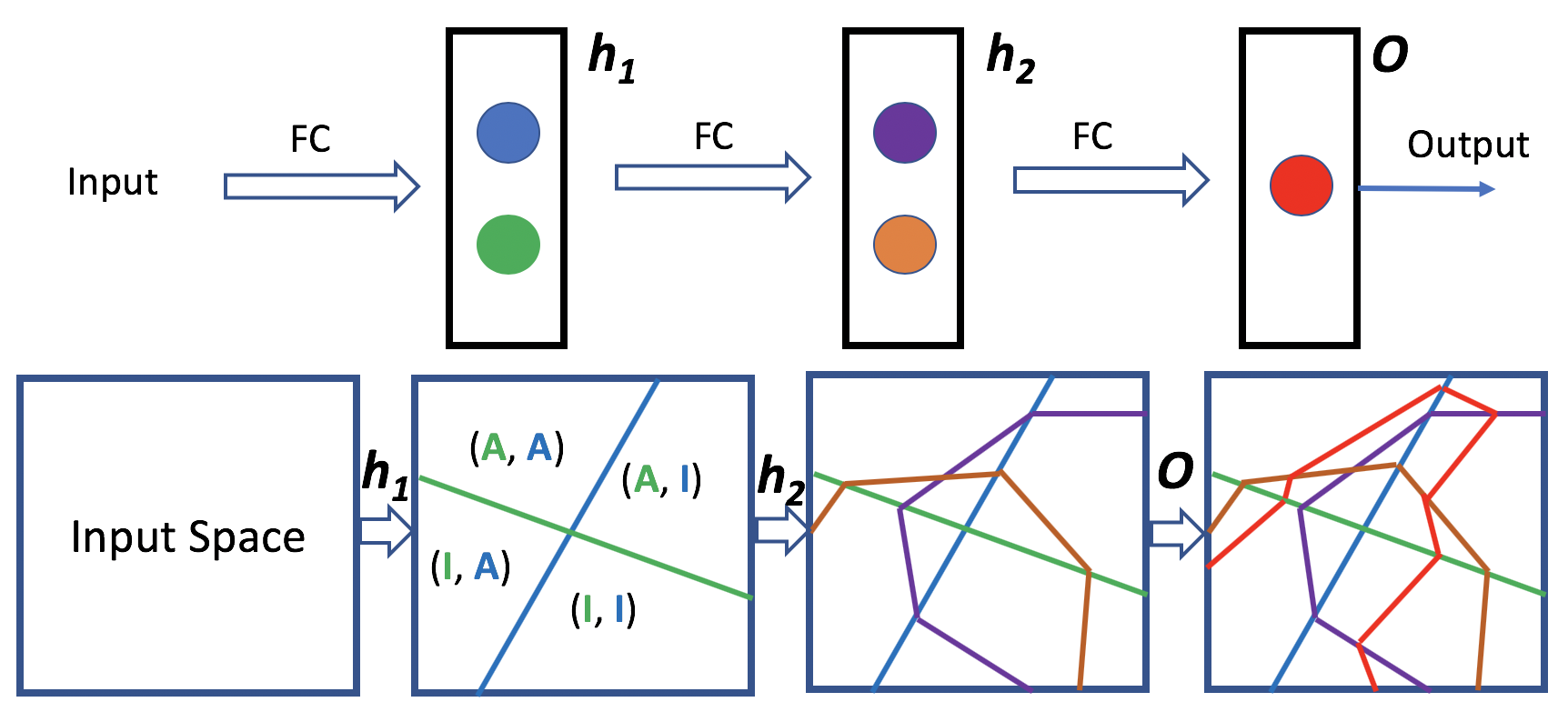} 
\caption{Illustration of how a fully connected DNN carves out the input space. Each layer progressively sub-divides the convex polytopes carved out by previous layers using bend-lines. This illustration is for general carving mechanism in any $N$-dimensional space which emphasizes that the linear function changes from polytope to polytope.}
\label{fig:carving_multilayer}
\end{figure}

\begin{figure}[ht]
\centering
\includegraphics[width=0.3\linewidth]{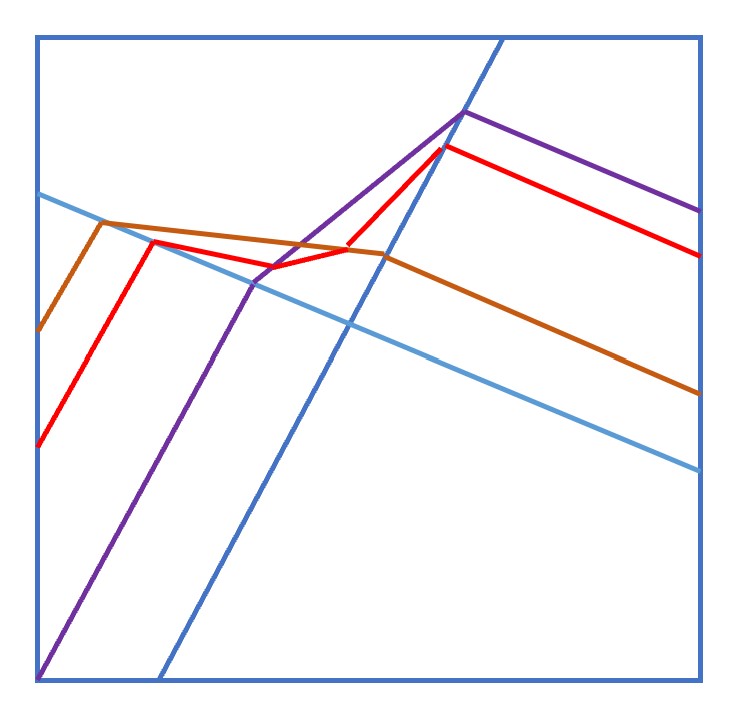} 
\caption{For an actual $N=2$ case, the carving will be typically simpler.}
\label{fig:carving_2d}
\end{figure}

\section{Piecewise Linear Function}
\label{sec:piecewise}
\index{piecewise linear function}
Let us unravel the answer to the question of how a neuron subdivides spaces which are carved by neurons in the layers preceding it.

\begin{figure}[ht]
\centering
\includegraphics[width=0.7\linewidth]{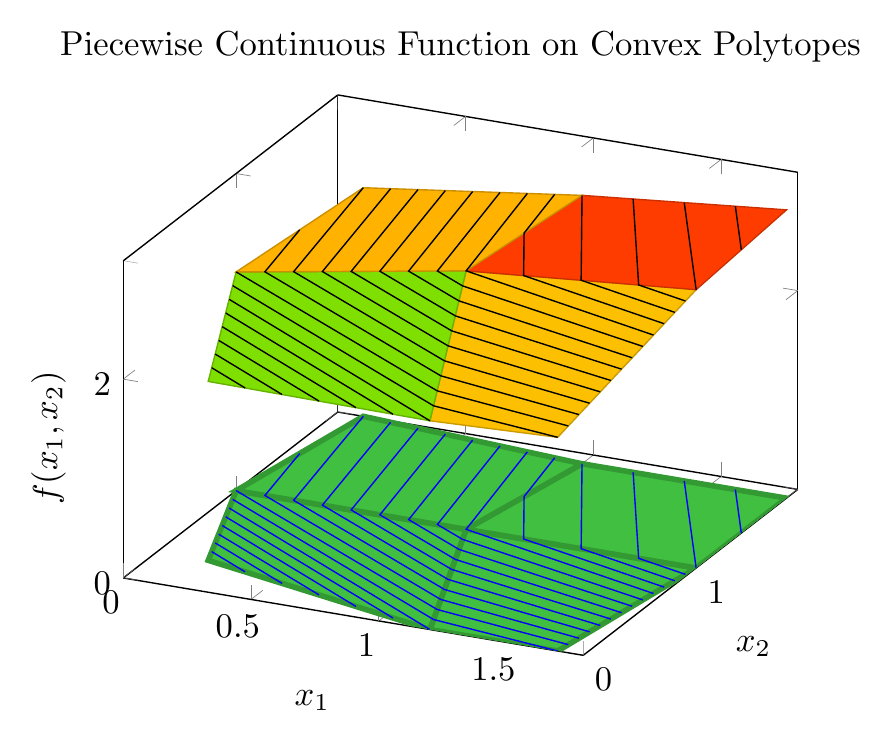} 
\caption{A piecewise continuous linear function fitted over convex polytopes. The parallel lines show the contour plot of the function for different values. A linear function has parallel hyperplanes as its contours. The gap between the parallel lines will be the same for a fixed stepsize of contour value.}
\label{fig:piecewise_linear}
\end{figure}

\subsection{Subdivision of the Input Space}
See Figure~\ref{fig:carving_multilayer} to develop intuition. In this illustrative example, the input is 2-D. There are two hidden layers $h_1$ and $h_2$, each with two neurons, and one output layer $O$ with a single output neuron.

Hidden layer $h_1$ divides up the input space into four convex polytopes with two lines corresponding to the two neurons. The activation pattern of each polytope is shown in parentheses. For example, the top right polytope has the activation pattern (A,I), which stands for (Active, Inactive) pattern. Inside this polytope, the green neuron is active and the blue neuron is inactive.

Consider the purple neuron in hidden layer $h_2$. Fix the activation pattern of neurons in hidden layer $h_1$ to (A,I). The purple neuron cuts the convex polytope with activation pattern (A,I) with a line, as shown. It is a linear function as the computation graph of the neural network does not have any non-linearity when you fix the activation pattern. Awake neurons become pass-through neurons and asleep neurons don't do anything. It is a composition of linear functions, which is again linear. Weights get multiplied and bias terms get added along the active computation paths, which start at the input layer and end at the output.

Move along the purple line from right to left. As soon as we step into the adjacent convex polytope with activation pattern (A,A), the blue neuron wakes up and gets activated. The linear function changes accordingly. It is a new composition of linear functions and the line bends. The change is continuous at the boundary of the two polytopes. We call the bending lines as bend-lines.
Similarly, the orange neuron of layer $h_2$ carves up the input space with another bend-line.

In total, we now have thirteen convex polytopes. Each polytope has a 4-D activation pattern
\[(a_1, b_1, a_2, b_2),\]
where $a_1, b_1$ are the two neurons in hidden layer $h_1$ and $a_2, b_2$ are the two neurons in hidden layer $h_2$.

The process continues to the output layer $O$. The red output neuron further carves the input space with a bend-line as it moves from one polytope to another. Follow the red line in Figure~\ref{fig:carving_multilayer} as it bends around in the space.

In total, we now have 21 polytopes. Each polytope has a 5-D activation pattern
\[(a_1, b_1, a_2, b_2, o)\]
where $o$ is the output neuron.
Each polytope has a linear function fitted on it. The whole input space has been fitted with a piecewise continuous linear function, see Figure~\ref{fig:piecewise_linear} for an illustration.

Figure~\ref{fig:carving_multilayer} is for a general illustration for $N$-dimensional case. For $N=2$, the carving will be typically simpler. For example, for (I,I) activation region the functions will be zero as the activation region disconnects the graph. For certain neurons and for certain polytopes, the lines will be parallel to the sides of the polytopes due to the constraints on the slope, see Figure~\ref{fig:carving_2d}.

The above analysis is for AI models which use ReLU as the activation function. This is true for most modern AI models in practice. Sometimes variations of ReLU functions\index{rectified linear unit!variations} such as ELU (Exponential Linear Unit), SELU (Scaled ELU), Leaky RELU, PReLU (Parametric ReLU) and Smooth-ReLU (also known as the softplus function\index{softplus function} $\ln(1 + \exp(z))$) are used.
What if the activation is a conventional non-linear function such as the sigmoid function or the hyperbolic tangent function? Then we lose piecewise linearity of the final function. Neurons will have values in the range $(0,1)$ and there are no on/off activation patterns of neurons. The result is a non-convex, non-linear function. The contour plot will consist of smooth curves. What has been observed is that ReLU-based AI models are much easier to train. It can be safely said that the use of the ReLU function contributed significantly to the success of deep learning, and that's why we have focused on the analysis of ReLU-based models.

\subsection{Piecewise Non-linear Function}
\label{sec:piecewise_nonlinear}
In classification, the piecewise linear function will get squished to the range $[0,1]$. The result is a piecewise non-linear function whose range is $[0,1]$. The contour plot still has parallel lines, but the function slopes non-linearly over each convex polytope.

Suppose the fitted function gives the probability $P(x)$ that the input $x$ is a cat. The final output is as follows.
$$
\text{Decision} =\begin{cases}
			\text{Cat}, & \text{if $P(x) > T_1$}\\
			\text{Dog}, & \text{if $P(x) < T_2$}\\
            \text{Indecision}, & \text{otherwise}
		 \end{cases}
$$
Thresholding at $T$ chops off a convex polytope along the contour line with value $T$. Therefore, each convex polytope will get sub-divided into at least one and at most three polytopes, corresponding to Cat, Dog and Indecision outcomes, see Figure~\ref{fig:convex_polytopes}.

\begin{figure}[ht]
\centering
\includegraphics[width=0.7\linewidth]{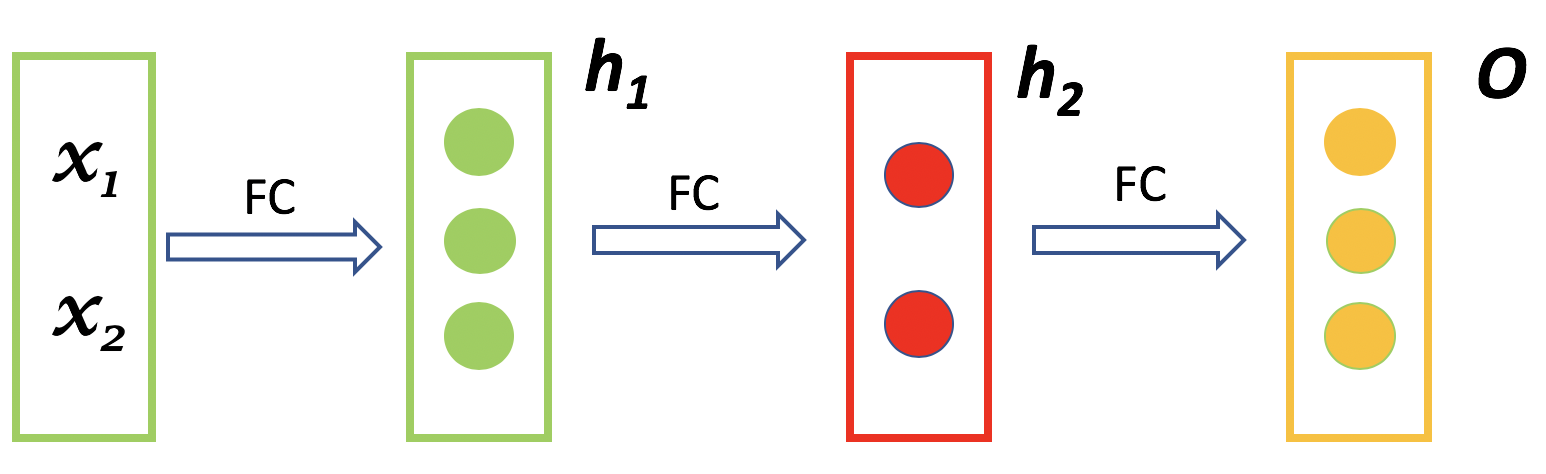} 
\caption{An example FFN with FC (fully connected) layers. See Figure~\ref{fig:full_example_carving} to see how it works.}
\label{fig:full_example_ffn}
\end{figure}

\begin{figure}[ht]
\centering
\includegraphics[width=\linewidth]{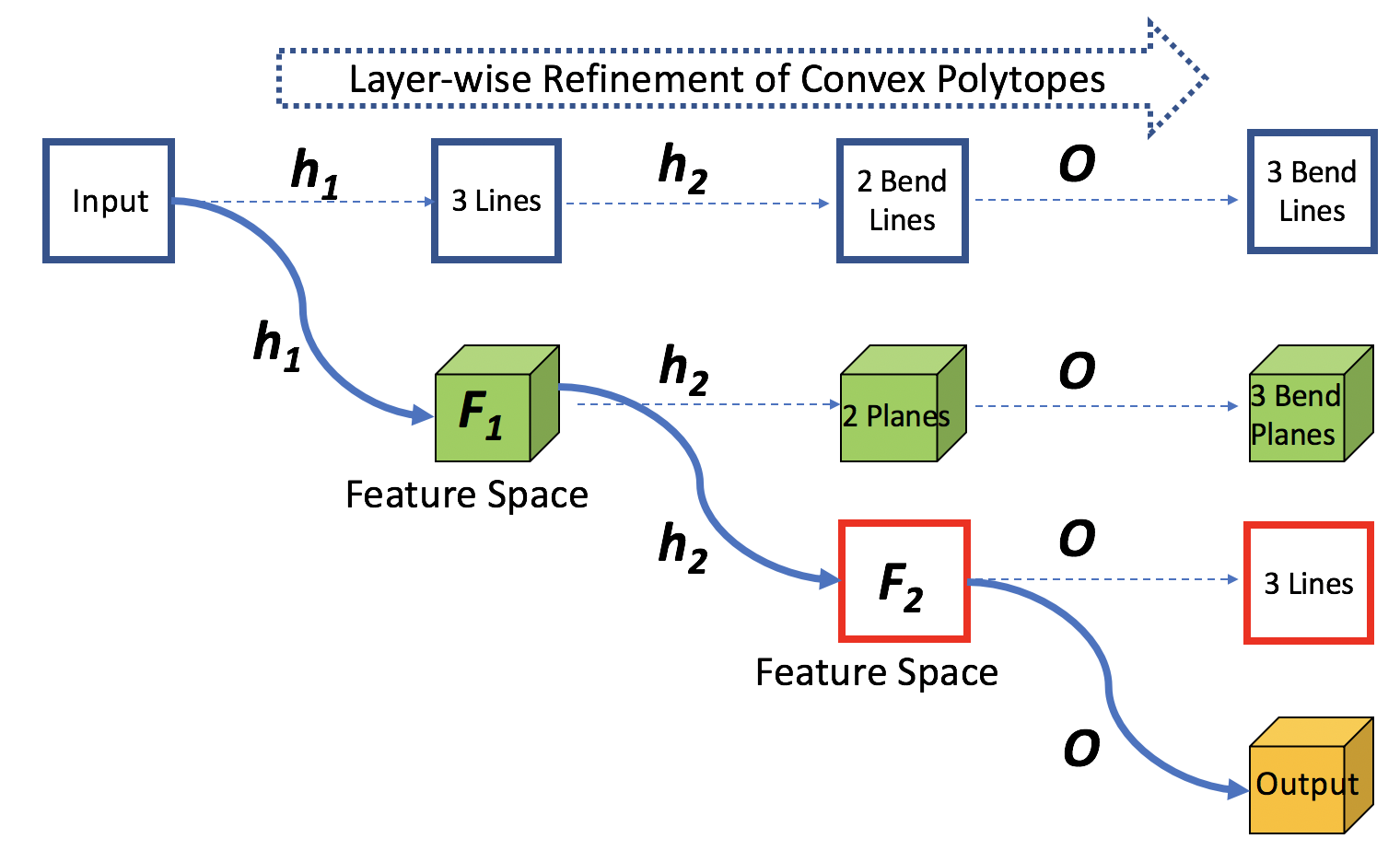} 
\caption{Carving of the input space and the feature spaces of the FFN shown in Figure~\ref{fig:full_example_ffn} by subdividing them into convex polytopes using bend-lines and bend-planes. Every layer $L$ is performing a simple carving of the preceding layer $L-1$, a slightly more complex carving of the layer $L-2$, and so forth. The solid blue curved arrows show the forward pass. Each row corresponds to a layer. Similarly each column corresponds to a layer. For example, the first row (the input layer) and the fourth column (the output layer $O$) show how the output layer carves the input space.}
\label{fig:full_example_carving}
\end{figure}

\subsection{Carving Out the Feature Spaces}
\label{sec:full_example_carving}
We now build a complete picture. See Figures~\ref{fig:full_example_ffn} and~\ref{fig:full_example_carving}.

In Figure~\ref{fig:full_example_ffn} we show a FC-FFN with two inputs. There are two hidden layers $h_1$ and $h_2$ with three and two neurons respectively. The output layer $O$ has three neurons. Each layer is a FC (fully connected) layer. We dive deeper into how each layer divides all the feature spaces (blobs) built by the preceding layers into convex polytopes.

See Figure~\ref{fig:full_example_carving}. The 2-D input space is transformed by layer $h_1$ into a 3-D feature blob $F_1$, which is again transformed by $h_2$ into a 2-D  blob $F_2$. Finally, the output layer takes us into a 3-D space. The output of a layer with $k$ neurons is a $k$-dimensional feature space.

Layer $h_1$ cuts up the input space with three lines. Layer $h_2$ divides the feature blob $F_1$ with two planes and the input space by two bend-lines. Finally, the output layer $O$ divides the feature blob $F_2$ by three lines, $F_1$ by three bend-planes and the input space by three bend-lines.

Every neuron in a layer $L$ is fitting a simple ReLU-activated linear function to the feature space spanned by the neurons in the preceding layer $L-1$, a slightly more complex piecewise linear function to the feature space spanned by the neurons in the layer $L-2$, and so forth, till it fits the most complex function to the input layer, see Figure~\ref{fig:full_example_carving}. Each neuron in a higher layer is trying to help separate the classes in the preceding feature space by computing a useful feature, which is a `feature of features' and a `pattern of patterns' and therefore it is a higher-level feature. It will be hard for a neuron in the first layer to do so because of complex geometry of manifolds, for which we need to bend around the manifolds.

Suppose we move along a straight line segment in the input space. What does this linear trajectory map to in the feature spaces computed by the hidden layers? Consider the activation pattern of the hidden layer $h_1$,
\[(a_1, a_2, a_3)\]
and suppose at the starting point of the line segment, it is
\begin{center}
    (A, A, A),
\end{center}
all three neurons activated. When we move in a straight line in the input space, then in the 3-D feature blob $F_1$, we move in a straight line because affine transformations map lines to lines. Note that the bias terms introduce translation making the transformations affine, a co-linearity preserving super-class of linear transformations. Suppose the activation pattern changes to
\begin{center}
    (A, A, I),
\end{center}
with one neuron getting inactivated. Now the trajectory in $F_1$ bends and then we move in another straight line in the 2-D plane spanned by the first two dimensions and with the last dimension being zero. If the three dimensions of $F_1$ are $(x,y,z)$, we are now moving in the $(x,y)$ plane. Suppose the activation pattern changes to
\begin{center}
    (A, I, I),
\end{center}
with the second neuron also getting inactivated. We are now moving along the $x$-axis. Therefore, a straight line in the input space bends around in the feature space.

\section{Expressive Power of AI}
\index{convex polytope}\index{linear region}\index{expressive power}
\label{sec:count_polytopes}
This section aims to develop an advanced intuition into why the expressive power of AI models is exponential in the depth of the models. The result is intuitive based on the geometrical carving process we have explained in the previous sections.

The goal is to find the bounds on the maximum realizable expressive power. A result on the upper bound has to show that there does not exist any weight assignment which will make a network exceed that bound. A result on the lower bound has to show that there exists some weight assignment which will make a network reach that expressive power.

We outline a high-level sketch of a proof that the number of convex polytopes (linear regions) is exponential in the depth of an AI model. Suppose the input is $d$-dimensional. We will keep $d$ fixed. Consider the first hidden layer with $n_1$ neurons. Each neuron cuts the $d$-dimensional space with a hyperplane. The total number of convex polytopes $r(n_1,d)$ carved by the first hidden layer of $n_1$ neurons in the $d$-dimensional input space is given by a beautiful result from the theory of hyperplane arrangements\index{hyperplane arrangement},
\[ r(n_1, d) = 1 + n_1 + \binom{n_1}{2} + \binom{n_1}{3} + \ldots + \binom{n_1}{d}, \]
see~\cite{HyperplaneStanley2007AnIT} for details.
Suppose the second hidden layer has $n_2$ neurons. Each $d$-dimensional polytope gets further subdivided by $n_2$ hyperplanes into $r(n_2,d)$ pieces, which bend around from one polytope to another. Therefore, the total number is given by
\[ r(n_1, d) r(n_2, d).\]
Of course, this is the most expressive carving which is possible and which is also desirable. If some bending hyperplanes don't intersect with some polytopes, then the number will be less. Hopefully, over training epochs the carving hyperplanes will shift around to increase the expressive power of the model.

For an example, refer to Figure~\ref{fig:convex_polytopes}. The first layer has three neurons, therefore we can confirm that
\[ r(3,2) = 7,\]
by looking at the straight dashed lines. The next processing layer, which is the output layer, has one neuron and it subdivides each of the seven polytopes into two parts with the bending dashed line, since $r(1,2) = 2$. In total, we have 14 polytopes. The reader should verify that if we had another neuron in this layer, then each of the seven polytopes could have been subdivided maximally into $r(2,2) = 4$ polytopes, yielding 28 polytopes in total. For another example, see Figure~\ref{fig:carving_multilayer}. The first hidden layer leads to $r(2,2) = 4$ polytopes. Each neuron in the second layer divides each of these four polytopes into two parts, but together they achieve maximal subdivision into four parts in only one of the polytopes.

Note that each term is a partial sum of binomial coefficients. There is no known closed formula for the partial sum of the first $d$ binomial coefficients. Note that, for fixed $d$,
\[ \binom{n}{d} = \Theta(n^d),\]
and therefore we have
\[ r(n, d) = \Theta(n^d),\]
that is, the function grows polynomially in $n$.

The theta ($\Theta$) notation used in computer science signifies both upper bound and lower bound. The omega ($\Omega$) and the big $O$ notation are for lower bound and upper bound, respectively. 
See~\cite{SimantBook} for the following result,
\[ r(n, d) = \Theta(n^d).\]

For two layers, we have the following upper bound on the maximal expressive power,
\[ O(n_1^d n_2^d).\]
Therefore if there is only one hidden layer then the number of convex polytopes grows polynomially with its width $n$ for a fixed $d$. If there are $L$ processing layers (including the output layer), we have
\[ O(n_1^d \ldots n_L^d),\]
and assuming that all layers have the same number $n$ of neurons, we have the upper bound
\[ O(n^{dL}),\]
which is exponential in $L$. Since $n$, $d$ and $L$ are typically large, an AI model has an astronomically high expressive power.
This is a remarkable result. Because of its importance, we will call it a mathematical theorem. 

{\bf Carving Theorem:}\index{carving theorem} Suppose a feed-forward fully-connected $L$-layer deep network has $d$-dimensional input and its each layer has $n$ neurons. Then, it carves the input space with $O(n^{dL})$ convex polytopes.

See~\cite{pmlr-v70-raghu17a-expressive} for the same result.
For a detailed formal proof based on tropical geometry, see~\cite{pmlr-v80-zhang-tropical}.

Is the above upper bound asymptotically tight? Can one choose the parameters of a network in such a way that the exponential expressive power is realized? We can choose the learnable parameters of a network such that multiple input activation regions get mapped to the same region in a feature space computed by a hidden layer. This can be done by $L-1$ consecutive layers to create exponentially many preimages. When the feature space computed by the $(L-1)$-th layer gets carved by the final hidden layer, then the carving gets replicated in all the preimages in the input space. See~\cite{NIPS2014_5422_Montufar_linear} for this construction and for the lower bound $\Omega((n/d)^{d(L-1)}n^d)$, $n \geq d$, which is exponential in depth and polynomial in width. The division by $d$ is a result of the special hand construction of the model. The first hidden layer folds the $d$-dimensional input space such that $(n/d)^d$ regions map to the same output region. This folding process is done by the first $L-1$ layers.

\section{Convolutional Neural Network}
\label{sec:carving_cnn}
\begin{figure}[ht]
\centering
\includegraphics[width=\linewidth]{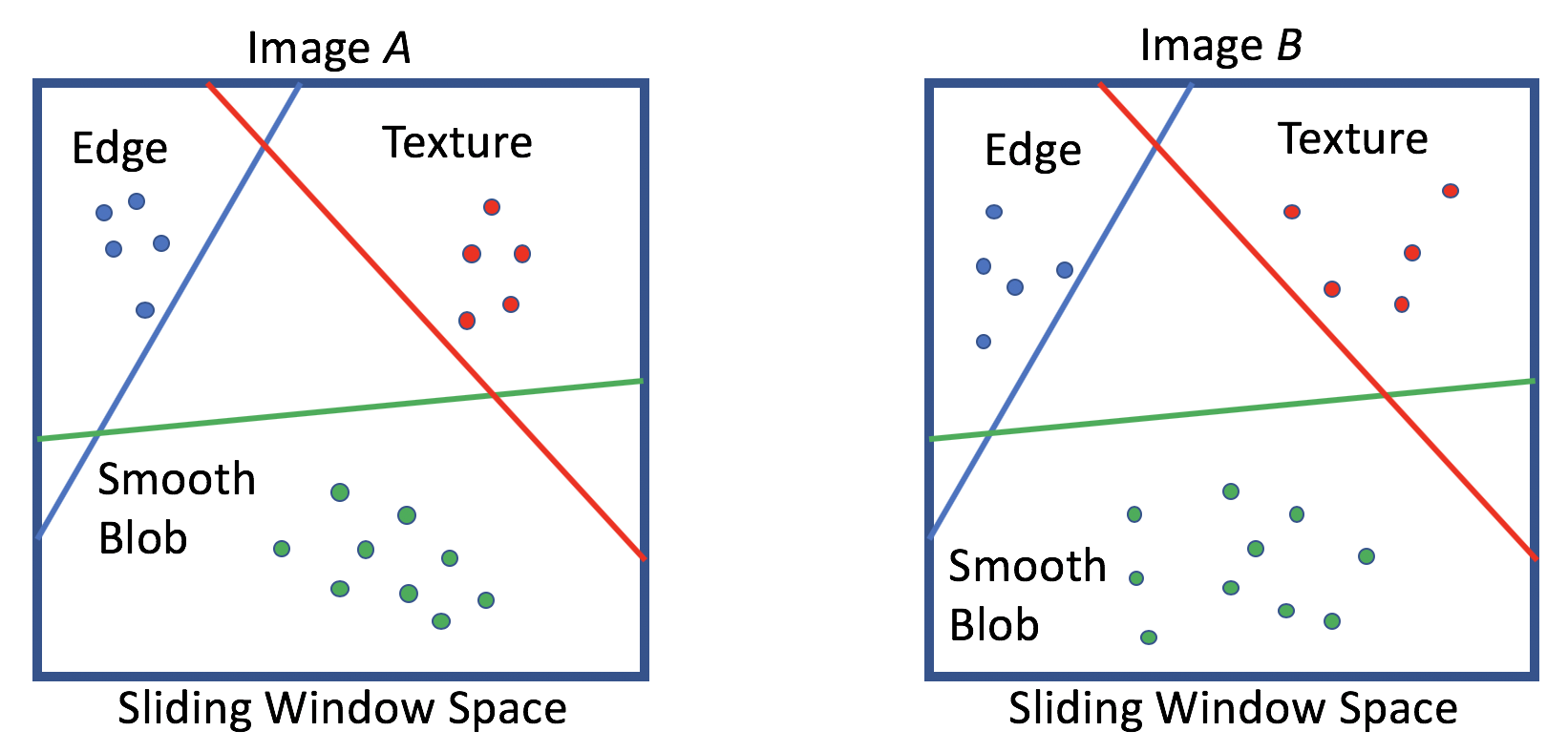} 
\caption{Carving up of the sliding window space by a CNN. Say for an image, an $n\times n$ convolutional filter is applied at $k$ positions. Sliding window space will have $k$ points in a $d = n\times n$-dimensional space.}
\label{fig:carving_cnn}
\end{figure}

\index{convolutional neural network}
We now develop an intuition into how CNNs carve up the input space. See Figure~\ref{fig:carving_cnn}. We will focus on the very first convolutional layer.

The same convolutional filter of dimension $n \times n$ is applied in a sliding window fashion across the image. Therefore, the input image has been divided into a set $X$ of $n \times n$ overlapping smaller images. Call $X$ the \emph{sliding window input space}. Convolutional filter operates like a fully connected hidden layer on the sliding window space.

In Figure~\ref{fig:carving_cnn}, we show the sliding window input spaces of two images $A$ and $B$. Each image yields $k$ many sliding windows. Therefore there are $k$ points in this $n \times n$ dimensional space. Suppose some of these are basic edge-like patterns from a visual perspective, some are textured, and remaining are smooth blob type.

Suppose we have three $n \times n$ convolutional filters. During training, assume that one filter gets trained to detect edge patterns, the second filter for textured patterns, and the third filter for blobs.
See Figure~\ref{fig:carving_cnn}, where on the left side, for image $A$, each point is $d=n\times n$-dimensional and represents an  $n \times n$ sliding window region. We have shown points of the three kinds (edge, texture, blob) and the corresponding convolution filter hyperplanes. For image $B$, these patterns can be in totally different spatial positions in the image. But due to translation invariance, the sliding window space of $B$ will be very similar to that of $A$, see Figure~\ref{fig:carving_cnn}.

Therefore, the convolutional filters lead to translation invariance in pattern matching.

For higher-level convolutional layers, the argument generalizes. We work with larger sliding window spaces of images, the dimensionality of which is given by the size of the receptive fields of the neurons. Instead of simple local patterns, convolutional filters now specialize for more complex patterns, which can be viewed as `patterns of patterns'. For example, a complex pattern may consist of a vertical edge on the left, a texture at the top, and a smooth green blob in the middle. Such higher-level patterns will get clustered in the sliding window space and a learned convolution filter will detect them. The underlying mechanism shown in Figure~\ref{fig:carving_cnn} remains the same. Starting from the local patterns, a CNN detects eyes, ears, tail and other parts of a cat, and finally the whole cat. The manifolds of image classes have a compositional topology and a CNN carves them out in a compositional manner.

We can understand the max pooling layer\index{max pooling} using the same intuition. Say it is a $2 \times 2$ max pooling layer. In the sliding window space, we have four points for a $2 \times 2$ pooling window. The point corresponding to the highest function value will be retained and the other three will be discarded. The highest value means that the point is furthest from the convolutional hyperplane in Figure~\ref{fig:carving_cnn}. The points closer to the border will get discarded. Therefore, there is immunity to local perturbations in the sliding window space.

\section{Recurrent Neural Network}
\label{sec:carving_rnn}\index{recurrent neural network}
We now develop an intuition into how RNNs carve up the input space.

\begin{figure}[ht]
\centering
\includegraphics[width=0.8\linewidth]{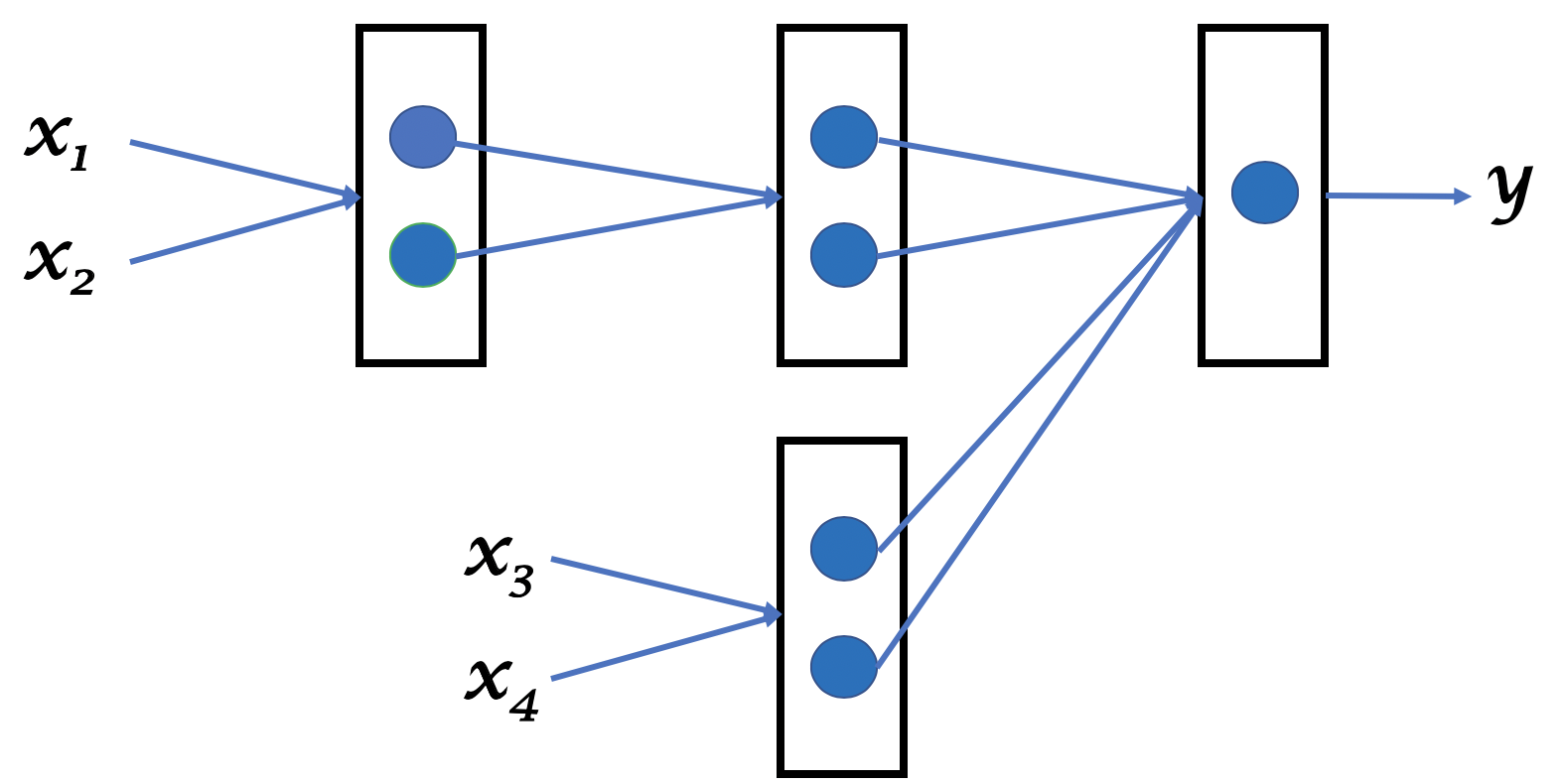} 
\caption{The convex polytope carving mechanism generalizes to directed acyclic computation graphs and to RNNs.}
\label{fig:carving_rnn}
\end{figure}

An RNN can be unfolded in time to turn it into a directed acyclic computation graph\index{computation graph}. See Figure~\ref{fig:carving_rnn} for an illustrative example. The input at the current time step is $(x_3,x_4)$. At the previous time step, the input was $(x_1, x_2)$. The current output  uses the current input and the previous hidden state.

Clearly, this is a special case of FFN with weight-sharing\index{weight sharing}. The mechanism of carving up the input space with convex polytopes and fitting them with piecewise linear functions holds fine.

Note that the input space is 4-D. One may ask how the output neuron carves up the 2-D subspace $(x_1,x_2)$. For that, fix some values of $x_3$ and $x_4$. Then, we get a 2-D cross section of the 4-D space. A cross section of a 4-D convex polytope is a convex polytope. Changing the values of $x_3$ and $x_4$ will effectively change an additive term into the output neuron and the corresponding cutting bend-hyperplane in the 2-D input subspace $(x_1,x_2)$ will shift.

\section{Attention and Carving}
\label{section:attention_carving}
\begin{figure}[ht]
\centering
\includegraphics[width=\linewidth]{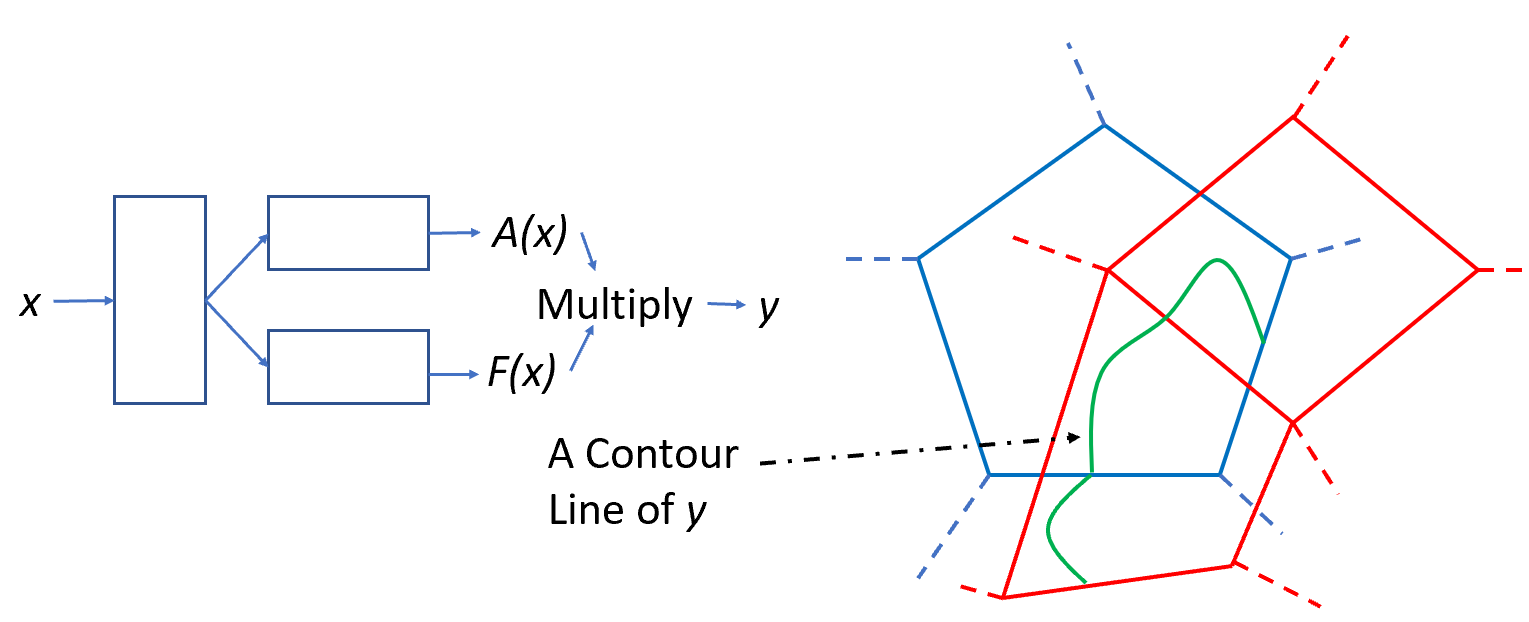} 
\caption{The attention mechanism involves modulating a feature with another through the multiplication operation. $A(x)$ and $F(x)$ are piecewise linear functions on their respective set of convex polytopes, shown in red and blue, respectively. Their product is a piecewise quadratic function where pieces are convex polytopes which are obtained by the intersection of the red and blue polytopes. A contour line in such a piece will be a conic curve in two dimensions. In three dimensions, it will be a conic surface. This contour line will bend when the activation pattern changes and it enters an adjacent piece.}
\label{fig:attention_carving}
\end{figure}

The attention mechanism\index{attention mechanism} modulates a feature by a weight which is dependent on the input. Therefore, we have a multiplication
\[y = A(x) F(x)\]
where the attention weight $A(x)$ is dynamically computed based on the specific input and the feature $F(x)$ is accordingly modulated. If $A(x)$ and $F(x)$ are continuous piecewise linear functions, then their product is a continuous piecewise quadratic function. Suppose the set of convex polytopes for $A(x)$ is $S_A$ and for $F(x)$ is $S_B$. The product $A(x)F(x)$ creates a piecewise quadratic function on a new set of convex polytopes $S$ obtained by intersecting polytopes in $S_A$ and $S_B$, see Figure~\ref{fig:attention_carving}. Each polytope in $S$ has an associated activation pattern of all neurons computing $A$ or $F$ considered together. Consider another example,
\begin{align*}
    & a = F_1(x) G_1(x) + F_2(x) G_2(x) + F_3(x) G_3(x),\\
    & y = [aF_1(x), aF_2(x), aF_3(x)].
\end{align*}
If $F$s and $G$s are ReLU networks, then $a$ is a piecewise quadratic function and each component of $y$ is a piecewise cubic function. Such dot product operations occur in attention networks, capsule networks\index{capsule neural network} and image similarity networks.

For each polytope in the intersection, the function changes quadratically. This significantly increases the non-linearity of the overall function being carved out by the AI model. We no longer have the property of straight lines being mapped to bending straight lines between the input space and the feature spaces. Now a line in one space maps to a non-linear curve in the other.

Suppose $y= f(x)$ is a quadratic function. The set of points in the input space for which $y = c$ for any constant $c$ is a conic surface, which is its level set. In a 2-D input space, it is a contour line of $y$. On one side of the curve, $y > c$ and on the other $y < c$. When the activation pattern changes, we cross over to an adjacent polytope and the conic changes, see Figure~\ref{fig:attention_carving}. It is a bending conic. We no longer have convex polytopes as pieces with flat faces on all sides. New added faces will be conic surfaces. Neurons in the layer preceding to the attention multiplication perform carving in the input space with bending hyperplanes. Neurons in the following layers do it with bending conics.

In self-attention\index{self-attention} used by transformers\index{transformer}, the same process is happening to a higher degree. What if we were to repeat it again? Consider two such steps of attention,
\begin{align*}
    y_1 & = A_1(x) F_1(x),\\
    y_2 & = A_2(y_1) F_2(y_1),
\end{align*}
where $A_1,A_2,F_1,F_2$ are the standard feed-forward networks and therefore they compute piecewise linear functions of their input. Since $y_1$ is a piecewise quadratic function, its composition with piecewise linear functions keeps $A_2(y_1)$ and $F_2(y_1)$ quadratic. The second multiplication is a product of two piecewise quadratic functions, making it a piecewise polynomial of degree four. The pieces correspond to different activation patterns. Note that pieces are mathematical objects with the level sets of polynomials as faces. The neurons subsequent to $y_2$ will perform carving of the input space with the bending level sets of polynomials of degree four. A convex polytope is a special case when the polynomial is of degree one and the level set is a flat face.

Note that in regular capsule networks and self-attention networks, there are additional non-linearities such as SoftMax in the middle of networks. Therefore, in the above example, we may have additional steps which process $y_1$. This makes the function more general enabling the model to perform more complex carving. The results in this section can be easily generalized to arbitrary non-linear functions and their level sets. We are ignoring these additional non-linear functions for simplicity and considering ReLU networks in which neurons can also multiply. Section~\ref{sec:piecewise_nonlinear} shows how Sigmoid or SoftMax introduces non-linearity. Instead of SoftMax, there can be other functions. It is an active research topic to find alternatives to SoftMax function in self-attention layers~\cite{performer} or to even eliminate it completely~\cite{richter2020normalized}.

We can generalize the carving theorem to include the multiplication neurons. 

{\bf Carving Theorem 2:}\index{carving theorem} In a feed-forward ReLU network, in which some neurons multiply the outputs of other neurons, a neuron computes a continuous piecewise polynomial function where pieces have level-sets of polynomials as their faces.

For a given hyperplane arrangement, if some of the hyperplanes are transformed to curves, that may potentially create even more complex arrangement of a larger number of regions, because the curve may bend around and intersect again. Just consider the case in which a conic curve intersects a straight line. There can be now two intersection points. Therefore, it only helps to have the multiplication neurons.

The proof of the theorem is same as for the simpler linear version of the theorem. In a way, it is a proof by induction by building these functions layer by layer as we move from the input to the output. This provides the basis of the carving algorithm\index{carving algorithm} which we now write down for a general network.
\begin{enumerate}
    \item Denote by $x$ the input to the network.
    \item Perform topological sorting of the directed acyclic computation graph of the network. Let the sorted sequence be $S$.
    \item Visit neurons in $S$ in the sorted order. For each visited neuron $y$, do the following.
    \begin{itemize}
        \item For each activation pattern of the neurons preceding to $y$ in $S$, on which $y$ is dependent, perform the following.
        \begin{itemize}
            \item Construct the polynomial function $P(x)$ being computed by $y$ by adding, multiplying and composing polynomials of the active preceding neurons in a forward pass.
            \item Draw the level set $P(x)=0$ in the region of the input space for which the activation pattern holds. On $P(x) > 0$ side of the level set, the neuron $y$ is active. The other side is zeroed out by ReLU.
        \end{itemize}
    \end{itemize}
\end{enumerate}
The above algorithm provides intuition into how AI carves manifolds. We can consider other activation functions. For example, consider that ReLU has been replaced with the sigmoid function everywhere. ReLU discretizes the firing activity of neurons into different regions. For the sigmoid function, all neurons are active, though some may be saturated towards 0 or 1. There is a single piece and a single highly non-linear function. Right at the output neuron, a level set will be created by a cutoff threshold to carve out the manifold for making the final binarized classification decision. The number of activation regions is one.

We end this section with a comment on a connection of AI with mathematics. Let us revisit level sets of polynomials. Consider a 2-D input space. Suppose a neuron is computing a polynomial of degree $d$. If $d=1$, we have bending lines. If $d>1$, we have bending conics. Now consider a 3-D input space, where we have surfaces. For $d=1$, we have bending hyperplanes. For $d>1$, we have bending conic surfaces. In an $n$-dimensional input space, we have bending high-dimensional conic surfaces. A level set is a solution to a single polynomial equation for a fixed activation pattern. The equation can be turned into an inequality to yield activation or inactivation regions.

How do these conic surfaces form the pieces of the piecewise polynomial function?
Consider a neuron $y$. Fix the activation pattern $A$ of the $N$ preceding neurons on which $y$ is dependent. A piece $X$ is being subdivided by $y$. $X$ is a solution to a system of $N$ \emph{multiple} polynomial inequalities, one for each neuron in $A$. The `sides' of $X$ are the solutions when one of the inequalities becomes an equation. This happens when we cross over into an adjacent piece and a neuron in $A$ just starts waking up or falling asleep. The `edges' and `corners' of $X$ are the solutions when multiple neurons start flipping their state. The neuron $y$ adds a new polynomial equation $y=0$ to the system, dividing $X$ into two subsets $X_1$ and $X_2$, for which $y>0$ and $y<0$, respectively. Note that when we move into an adjacent piece, the polynomial system has to be recomputed for a changed activation pattern $A'$.

Solving a system of a large number of polynomial equations with a large number of variables is one of the great challenges of the twenty-first century mathematics. It is remarkable to see this connection between modern AI models and algebraic geometry. An open problem is to rework the bounds on the exponential expressive power for piecewise polynomial networks. And, what will happen if we add neurons for division of polynomials which leads to rational functions? We anticipate that Bayesian inference will play a role in the future of robust and interpretable AI. The probabilities computed by the AI models will get multiplied and divided by other probabilities. Such models will be understood at two levels. One will be at a high level in which we explicitly think in terms of a compositional process consisting of parts and whole. The other will be to understand how such methods will carve out highly complex manifolds. For additional examples of carving of manifolds by AI models, see~\cite{SimantBook}.

\begin{figure}
\centering
\begin{tikzpicture}[
plain/.style={
  draw=none,
  fill=none,
  },
net/.style={
  matrix of nodes,
  ultra thick,
  nodes={
    draw,
    blue,
    circle,
    inner sep=10pt
    },
  nodes in empty cells,
  column sep=1cm,
  row sep=-7pt
  },
>=latex
]
\matrix[net] (mat)
{
|[plain]| \parbox{1.3cm}{\centering Input\\Layer} & |[plain]| \parbox{1.3cm}{\centering Hidden\\Layer} & |[plain]| \parbox{1.3cm}{\centering Output\\Layer} \\
$x_1$ & |[plain]| \\
|[plain]| & $h_1$\\
$x_2$ & |[plain]| \\
  |[plain]| & |[plain]| \\
$x_3$ & $h_2$ & $y$ \\
  |[plain]| & |[plain]| \\
$x_4$ & |[plain]| \\
  |[plain]| & $h_3$ \\
$x_5$ & |[plain]| \\    };
\foreach \ai [count=\mi ]in {2,4,...,8, 10}
  \draw[<-, purple, ultra thick] (mat-\ai-1) -- node[above] {$x_\mi$} +(-2.7cm,0);
\foreach \ai in {2,4,...,8, 10}
{\foreach \aii in {3,6,9}
  \draw[->, blue, thick] (mat-\ai-1) -- (mat-\aii-2);
}
\foreach \ai in {3,6,9}
  \draw[->, blue, thick] (mat-\ai-2) -- (mat-6-3);
\draw[->, purple, ultra thick] (mat-6-3) -- node[above] {Output} +(3.5cm,0);

\end{tikzpicture}
\caption{A fully connected FFN with one hidden layer.}
\label{fig:one_layer_ffn}
\end{figure}
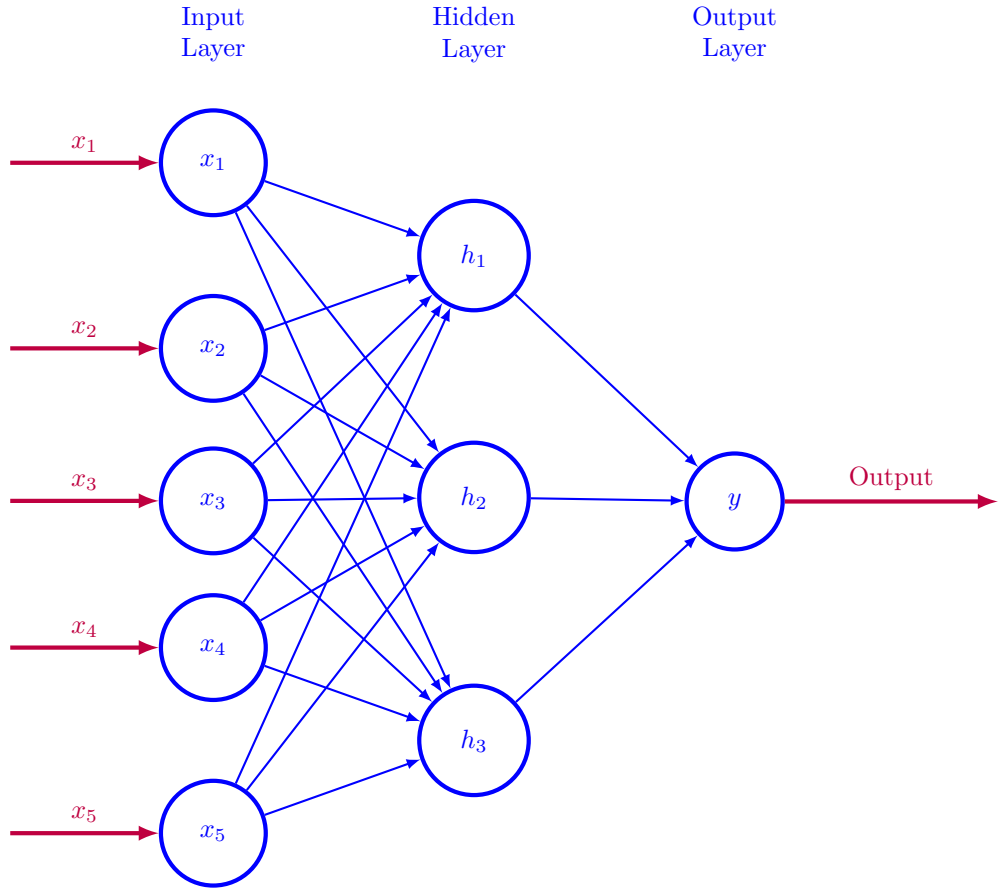

\section{Optimization Landscape}
\index{optimization landscape}\index{loss landscape}
\subsection{Graph-Induced Polynomial}
See Figure~\ref{fig:one_layer_ffn}. There are 15 computation paths connecting an input to the output. As an example, 
\[x_3 \rightarrow h_1 \rightarrow y \]
is one of these paths. Suppose the weight parameter of the edge from $x_3$ to $h_1$ is $w_{3,1}$ and of the edge from $h_1$ to $y$ is $u_1$. Then, if $h_1$ is active, the input dependent term
\[x_3 w_{3,1} u_1\]
will contribute a term in the computation of the output. Each path starting from an input will contribute such a term. If the bias parameter for $h_1$ is $b_1$ then the bias term
$b_1 u_1$
will be computed which is input independent. Each path starting from a bias parameter will contribute such a term. There are 4 such paths. All these terms combine to form a polynomial.

We have seen that the output of a ReLU AI model is a continuous piecewise linear function of the input, where the linear coefficients are given by the activated part of the  computation graph of the neural network based on the activation pattern invoked for the given $N$-dimensional input,
\[ y = f(x) =  \alpha_0 + \sum_{i=1}^N \alpha_i x_i.\]
The linear coefficient $\alpha_i$ for the input $x_i$ will be the sum of the products of the weights over all the activated paths in the computation graph from the input $x_i$ to the output $y$. If we were to keep $x_i$'s constant, and vary the weights and biases, then
\[y = f(\theta)\]
is a polynomial in learnable parameters, where $y$ is the output and $\theta$ is the set of learnable parameters. For classification, there is the additional step of squishing the output to the range $[0,1]$. For regression, $y$ is the answer.

The degree of a monomial term in the polynomial is the length of the path, and since weights are typically not shared between edges along a computation path, it is a multi-linear term of the form
\[ x_i w_{i_1} \ldots w_{i_d},\]
where the path of length $d$ starts at the input $x_i$, and the maximum degree of a weight term is 1. Each monomial term has degree equal to or less than the depth of the AI model. For networks with the attention mechanism, we don't have multi-linear terms because the same weight may be shared by the feature network and the attention network, and later in the network we will have a multiplication of the attention weight with the feature value.

In the previous section, the input was variable and the parameters were fixed. In this section, for a fixed input, we compute a high-degree polynomial of the learnable parameters. By composing the loss function and this polynomial, we get the loss surface over the parameter space.

\subsection{Gradient of the Loss Function}
\index{loss function}
The loss function depends on the output $y$ and the ground truth $G$. For a minibatch, we take the aggregate of the loss functions of several inputs. Note that for the $L^2$ loss, the landscape will be a polynomial
\[\mathcal{L}(\theta) = (f(\theta) - G)^2.\]
For any learnable weight $w \in \theta$,
\[ \frac{\partial \mathcal{L} (\theta)}{\partial w} =  2 (f(\theta) - G)  \frac{\partial f(\theta)}{\partial w},\]
therefore, the gradient will be zero if either $f(\theta) = G$ or if the gradient of polynomial $f(\theta)$ is zero.
For the cross-entropy loss,
\begin{align*}
        & p = \text{sigmoid}(f(\theta)) = 1/(1 + \exp(-f(\theta))), \\
        & \mathcal{L} (\theta) = - G \log(p) - (1-G) \log(1-p).
\end{align*}
Therefore, we have for any $w \in \theta$,
\[ \frac{\partial \mathcal{L}(\theta)}{\partial w} =  (p - G)  \frac{\partial f(\theta)}{\partial w}.\]
Therefore, the gradient will be zero if either $p = G$ or if the gradient of the polynomial $f(\theta)$ is zero.

\begin{figure}[!ht]
\centering
\begin{tikzpicture}
    \node[anchor=south west] (image) at (0,0) 
        {\includegraphics[width=\textwidth]{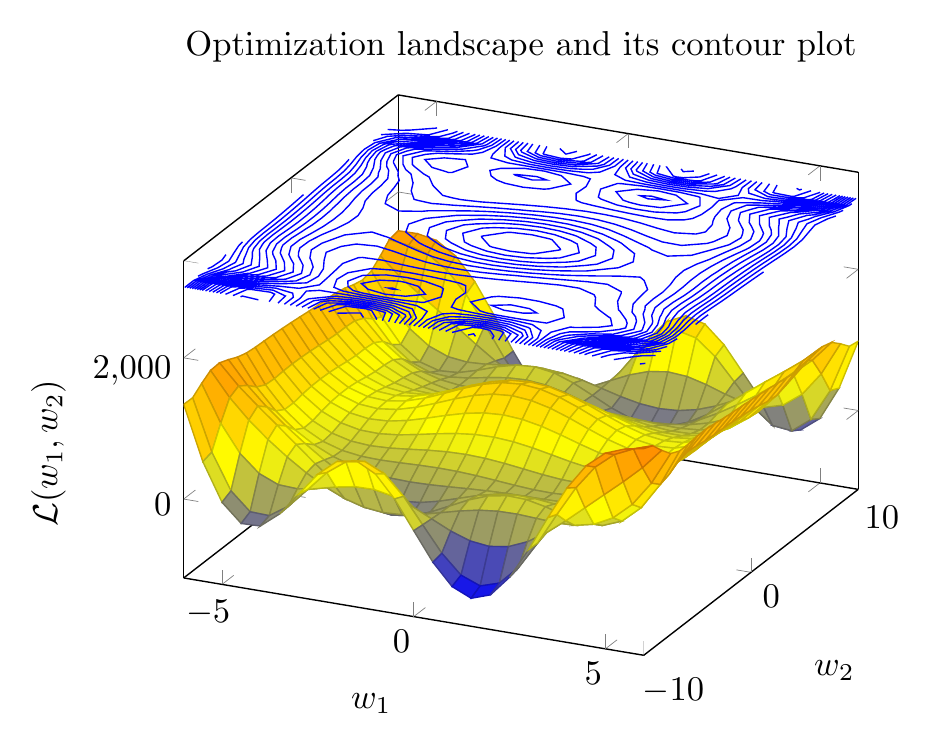}};
\end{tikzpicture}
\caption{The optimization landscape (loss surface) of a loss function where there are two learnable parameters. SGD computes gradients of the loss function with respect to the two parameters. The gradient is a 2-D vector on the contour plot and it will be perpendicular to the contour at that point.}
\label{fig:optimization_landscape_contour}
\end{figure}

\subsection{Visualization}
\index{optimization landscape!visualization}
See Figure~\ref{fig:optimization_landscape_contour}for examples of loss functions over a 2-D parameter space. You can see numerous hills and valleys. In Figure~\ref{fig:optimization_landscape_contour}, the contour plot is also shown. Imagine yourself following the SGD algorithm and sliding down slopes till you find yourself at the bottom of a valley.

\begin{figure}[ht]
\centering
\includegraphics[width=\linewidth]{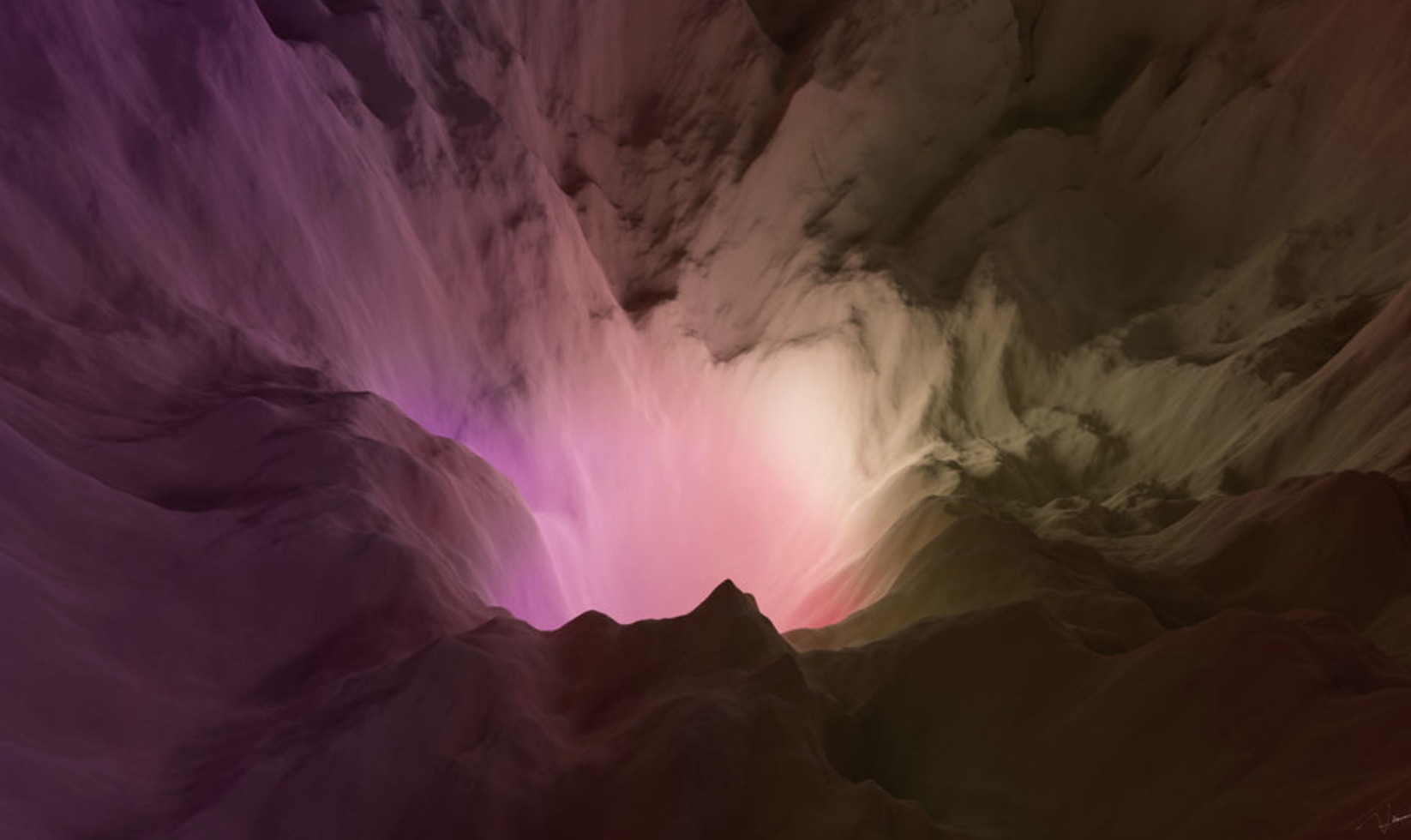} 
\caption{Optimization landscape of an actual loss function over a 2-D cross section of a high-dimensional space. This image of a loss landscape is created with data from the training process of a convolutional neural network. Specifications: Imagenette dataset, SGD-Adam, minibatch size of 16, batch normalization, learning rate scheduler, train mode, one million loss points, log scaled. This piece has been created through a multi-stage process that begins with the training of a deep learning network. Created by Javier Ideami. For more about the Loss Landscape project visit \url{https://losslandscape.com}.}
\label{fig:javier_losslandscape}
\end{figure}

AI does the same with the big difference that it works in a high-dimensional space. Say there are 10 million learnable parameters in a CNN, therefore it is a 10 million-dimensional space, which is way beyond our visualization capability.
In order to get a feel for the surreal landscape, one can choose a point in the trajectory of the SGD optimizer in this space and look at its vicinity. In particular, one looks at a planar cross section defined by two random, almost orthogonal, directions as the axes. The loss surface is plotted over this 2-D plane. See Figure~\ref{fig:javier_losslandscape}, which shows a result of this approach.
Such appealing visualizations of loss surfaces in 2-D lead to interesting insights into their nature, see~\cite{visualloss}. One can visually observe their rugged hills and valleys, and how different techniques such as dropout or batch normalization make the surface smoother or rougher. In fact, local minima at the bottom of the landscape seem to be connected by simple paths, see~\cite{Garipov2018LossSM}. It is like taking a walk from one crater to another along a flat smooth path in a valley on an alien planet. This suggests exploring the surrounding valley by creatively varying the learning rate and taking an ensemble of the nearby models corresponding to low error values in a method known as fast geometric ensembling~\cite{Garipov2018LossSM}. By taking a running average of weights of these models rather than their predictions, in a method called stochastic weight averaging~\cite{stochastic_weight_averaging}, one can approximate the ensemble by a single model. Applying this averaging during the entire SGD leads to less bumpy descent and smoother error curves.

Empirical experiments suggest that typical optimization landscapes have a simple geometrical structure. For example, if one were to plot the loss values along a line from the initial position $\theta_0$ to the final position $\theta_f$ in the parameter space, 
\[ \mathcal{L}( (1 - \alpha) \theta_0 + \alpha \theta_f),\]
by varying $\alpha$ from 0 to 1,
then one gets simple, approximately convex curves, see~\cite{Goodfellow2015_qualitatively_opt_land}. Another way to investigate optimization landscapes is by restricting the number of dimensions in which the parameters are allowed to change. That is, the gradient descent is forced to occur in a $d$-dimensional subspace whose basis is chosen randomly in the complete $D$-dimensional space. Empirically one can vary $d$, and find the transition when SGD starts finding good solutions. Interestingly, these experiments suggest that $d$ is not as large as one might expect. Furthermore, comparisons of this empirical intrinsic dimension\index{intrinsic dimension} of the landscape for different problems allow one to measure the relative difficulty levels of these problems, see~\cite{intrinsic_dim_obj_land_2018}.

\subsection{Critical Points}
\index{critical point}
Fortunately, a large number of dimensions means that there is almost always an opportunity to tweak some parameter(s) to continue the slide downwards. There are many degrees of freedom available to improve the AI model.
Recent empirical and theoretical work strongly suggests that local minima are relatively more concentrated towards the bottom of the optimization landscape whereas saddle points are more numerous away from the bottom. Furthermore, most critical points are saddle points. Therefore during training we are likely to encounter local minima towards the end which correspond to good solutions to practical problems. Earlier in the training, it is highly likely that temporary slowing down of the training is because of a saddle point. A saddle point in Figures~\ref{fig:optimization_landscape_contour} and \ref{fig:javier_losslandscape} will resemble a saddle on a horse. Gradients are zero but it is neither a minimum nor a maximum. There are directions in which it is a minimum and in others it is a maximum. Its graph crosses the tangent plane at the point.

The presence of saddle points provides a justification for stochastic algorithms. The stochastic nature of SGD\index{stochastic gradient descent} can assist in escaping saddle points by finding a direction which opens up on a new minibatch of data, that is, the gradient is non-zero in that direction and allows further descent towards the minimization of the loss function. The fluidity of the optimization landscape is helpful in escaping a saddle point.
Consider the scenario in which ReLU neurons wake up or go to sleep on new minibatch samples, thereby having ripple effects on the bend-hyperplanes of neurons of higher layers they are connected to. We find ourselves in new convex polytopes and the loss surface changes. This noisy nature of SGD would help in escaping from a critical point\index{critical point}.

And it is rare to encounter local minima.
The way AI carves up the input space in terms of convex polytopes comes to our rescue. A new minibatch represents a new set of points. The carving bend-hyperplanes have to adjust to separate out the classes. It is hard to get stuck in a local minimum because there is room to tweak some bend-hyperplanes, which are as numerous as the number of neurons.

In high-school calculus, we learned that a local minimum\index{local minima} of a function $f(x)$ satisfies:
\[f'(x) = 0, f''(x) > 0.\]
The first derivative has to be zero and the second derivative has to be positive.
For multi-variate functions we replace the conditions with those on vectors and matrices.
\begin{enumerate}
    \item The gradient vector has to be the zero vector for a critical point. The gradient vector is made of all first partial derivatives.
    \item For the test of the second derivative being positive for a local minimum, we now require the Hessian matrix\index{Hessian matrix} to be positive definite. The Hessian matrix is made of all second partial derivatives.
\end{enumerate}
That is, the eigenvalues\index{eigenvalue} of the Hessian matrix have to be all positive at a local minimum. If the Hessian matrix is negative definite, then the point is a local maxima. If the eigenvalues are both negative and positive, then it is a saddle point\index{saddle point}. If any eigenvalue is zero, it wipes out the determinant of the Hessian matrix, making it a singular matrix, and then the critical point can be anything --- local minimum, local maximum, saddle point or none of these. In univariate case, this corresponds to the case when $f''(x) = 0$.

For random matrices, one can ask what the probability of all eigenvalues being negative is. For a special class of random matrices, known as Gaussian orthogonal ensembles, it can be shown that the probability of a critical point being a local minimum is very small. This result relates to a similar one for random polynomials, whose critical points are overwhelmingly saddle points. Thus it is highly unlikely that SGD will get stuck in a local minimum.

Moreover, saddle points are easy to escape early in the training when the loss is high. Towards the bottom of the landscape when the AI model is working well, they become harder to escape. Let's develop some intuition into this claim.
Let index be the fraction of negative eigenvalues of the Hessian matrix. There is very little likelihood that all eigenvalues will be positive when the loss function is high because there are a large number of parameters to tweak. The index will be high. As the training proceeds, the index will reduce gradually and when it reaches a local minimum, then it will become zero. When the loss is high, high-index saddle points dominate. As we descend the landscape, low-index saddle points become relatively more numerous. Intuitively, high loss implies that there are several directions in which one can descend down and there are many degrees of freedom to fix the errors and improve the model. As one approaches the bottom of the landscape, loss is low, the model is working great, errors are few and there are few directions available to improve the model further. The optimization landscape has this interesting, very intuitive, layered structure. In addition, empirical results suggest that during and after training, most of eigen-values of the Hessian matrix are zero making it singular. This implies there is an overall flatness in the layered optimization landscape.

In physics, spin glasses\index{spin glass} and Ising models\index{Ising model} have been studied to model systems in which particles interact with others under constraints. Their critical points have the same structure because there are many degrees of freedom to lower the energy of such physical systems. This provides a connection between AI and existing results. See~\cite{Choromanska_loss_surface_2015, Dauphin_saddle_2014} to get an idea of how researchers are making progress in the analysis of optimization landscapes. We will continue to see more results in the future.

\section{The Mathematics of Loss Landscapes}
The previous section is a step in developing high-level intuition about how an AI model gets trained in a high-dimensional loss landscape. This section is written to build advanced intuition and greater mathematical insight. The mathematics needed to understand such landscapes is challenging, yet it is possible to develop basic conceptual understanding.

\subsection{Random Polynomial Perspective}
\index{random polynomial}
We saw that the output of a ReLU AI model is a polynomial $f(\theta)$ and the critical points of the loss function $\mathcal{L}(\theta)$ coincide with those of the polynomial due to the chain rule of differentiation. Therefore we can look at the nature of critical points of polynomials and ask what is the probability of a critical point being a local minimum.

We will look at a special class of polynomials under certain specific assumptions to get an idea about the critical points of large polynomials.
Consider a space of polynomials with degree two in two variables,
\[f(x,y) = a_1 x^2 + a_2 y^2 + a_3 x y + a_4 x + a_5 y + a_6,\]
where each $a_i$ is an independent, centered, random normal variable. The variances are chosen to be multinomial coefficients, where the central terms, which can come from multiple sources, have higher variance. This space can be generalized to a space of polynomials with degree $d$ in $n$ variables,
\[f(x) = \sum_{|\alpha| \leq d} f_\alpha x_1^{\alpha_1} x_2^{\alpha_2} \ldots  x_n^{\alpha_n},\]
where $\alpha = (\alpha_1, \ldots, \alpha_n) \in \mathbb{N}^n$ is a multi-index and $|\alpha| = \alpha_1 + \ldots + \alpha_n$. If we set the variances of the polynomial coefficients to the multinomial coefficients,
\[ \frac{d!}{\alpha_1! \ldots \alpha_n! (d - |\alpha|)! },\]
then we get the space of random normal polynomials. Note that the variances indicate the number of sources which can contribute to the monomial term. For degree $d$, there are at most $d$ sources, from each of which one can select a variable or choose not to select at all. Therefore, for the case $d=n=2$, the term $x^2$ will have half the variance of the term $xy$, because for the latter there are twice as many ways to create the monomial. For the random normal polynomial space, using the big $O$ notation from computer science, the probability that a critical point is a local minimum is
\[\Pr(\text{Local Minimum}) = O(\exp(-k n^2)) ,\]
where $k \approx 0.275$ (its exact value being $(\ln 3)/4$). This is quite amazing. The probability falls off exponentially as the number of learnable parameters $n$ increases, which can be in tens of millions. The expected number of critical points increases exponentially at the same time,
\[ O ( (d-1) ^ {(n+1)/2}),\]
with almost all being saddle points\index{saddle point}. Note that the degree $d$ is the depth of the AI model. See~\cite{random_polynomial_DEDIEU200889} for details.

\subsection{Random Matrix Perspective}
\label{subsection:random_matrix}
\index{random matrix}
The previous subsection presents the viewpoint from the random polynomial perspective. An alternative viewpoint is that of random matrices. In fact, the results about critical points of random polynomials are based on results from random matrix theory, in which real, symmetric, random, normal matrices called Gaussian orthogonal ensembles\index{Gaussian orthogonal ensemble} are used.

We mentioned the second-derivative matrix known as the Hessian matrix in the previous section. The eigenvalues\index{eigenvalue} of the Hessian matrix\index{Hessian matrix} at any point in the optimization landscape capture the geometry of the loss surface at that point in terms of convexity and concavity of the local patch. The $n \times n$ Hessian matrix of a function $f(x_1,\ldots,x_n)$ is
\[
  H_{ij} = \frac{\partial^{2} f}{\partial x_{i} \partial x_{j} }.
\]
In AI it is a symmetric matrix because second partial derivatives are continuous. Consider a 2-D optimization landscape. Let the eigenvalues at a given point be $\lambda_1$ and $\lambda_2$. If $\lambda_1 > 0, \lambda_2 > 0$, then the Hessian matrix is positive definite, and the local patch around that point is concave up. If  $\lambda_1 < 0, \lambda_2 < 0$, then the patch is convex down. If one is positive and the other is negative, it is concave up in one direction and convex down in another. If the point is a critical point, that is, the gradient is zero at that point, then the implication is that the critical point is a local minimum, a local maximum or a saddle point\index{saddle point}, respectively for the three cases. If any eigenvalue is zero, then we can't say much, and one really has to plot the function to draw conclusions.
For an example, suppose $f(x,y) = 2x^3 + y^3 - x y$. Then,
\[
  \text{Hessian} (f(x,y)) = 
  \begin{bmatrix}
    12x & -1 \\
    -1 & 6y
  \end{bmatrix}
.\]
Note that at point (0,0), the eigenvalues are $\pm 1$. Therefore, the patch at (0,0) is both convex down and concave up.

\begin{figure}[t]
\includegraphics[width=0.25\linewidth]{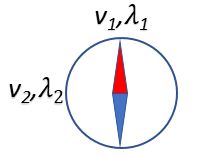} 
\caption{If you are lost in the loss landscape with very poor visibility, the best thing is to have a compass which shows the eigenvectors and corresponding eigenvalues. That will give you some idea about the geometry of the local patch. If at some point, you notice that all eigen-values in your eigen-compass are registering positive values, then it will mean you have gotten trapped in a local minimum. How likely is this?}
\label{fig:eigen_compass}
\end{figure}

In order to see how eigenvalues describe the geometry around a critical point, consider the local Taylor series approximation around it,
\[ f(x + \triangle x,y + \triangle y) = f(x, y) + \frac{1}{2} (\triangle x,\triangle y)^T H (\triangle x, \triangle y),\]
where $H$ is the Hessian matrix.  This is a second-order quadratic approximation of the landscape. Note that the first-order term drops out as the gradient is zero. Let $\lambda_1 > \lambda_2$ and let $v_1$ and $v_2$ be the corresponding orthonormal eigenvectors\index{eigenvector}. If the vector $(\triangle x,\triangle y)$ is $v_1$, then the difference
\[ f(x + \triangle x,y + \triangle y) - f(x,y) = \frac{1}{2} \lambda_1\]
is maximized, where we used the fact that
\[ H v_1 = \lambda_1 v_1\]
and that $v_1$ is of unit length.
By moving along in this direction, if $\lambda_1 > 0$, then we have the steepest ascent. Similarly, the loss will decrease along an eigen-vector with a negative eigenvalue. The sign of an eigenvalue determines whether moving along the corresponding eigenvector will be an ascent or a descent, and its magnitude determines by how much. See Figure~\ref{fig:eigen_compass}.

We want to know what the probability is of all eigenvalues being positive. For Gaussian orthogonal ensembles, the probability is
\[\Pr(\text{All Eigenvalues are Positive}) = O(\exp(-k n^2))\]
where $k \approx 0.275$, see~\cite{random_matrix_GOE_2006}. This result was
used to obtain the result for random normal polynomials in the previous subsection.
Gaussian orthogonal ensembles provide a model for the Hessian matrices, therefore it is highly unlikely that a critical point will be a local minimum. If we had modeled an eigenvalue being positive or negative as a Bernoulli process, like a coin toss, we would have concluded:
\[\Pr(\text{All Eigenvalues are Positive}) = O(2^{-n}).\]
This is exponential decay with $n$ but actual decay is significantly more rapid due to the squared term. Here is the intuition behind this amazing phenomenon. Coin tosses would have modeled the eigenvalues as independent random variables. In reality, they are dependent on each other in an intriguing way. A physical analogy is that of charged particles as in a Coulomb gas. The eigenvalues are restricted to a 1-D line and repel each other with a physical force. This repulsive force makes it highly unlikely that they will all get clustered on the positive side. In other words, local minima are exceedingly rare. SGD is likely to encounter only saddle points.

\subsection{Spin Glass Perspective}
\label{subsection:spin_glass}
\index{spin glass}
A number of systems in physics, computer science and mathematics are defined in terms of interacting components in which we have to minimize an objective function. These interactions make it difficult to satisfy all constraints which can conflict with each other. In physics, a spin glass or an Ising model\index{Ising model} consists of particles with couplings which indicate the mutual constraints. The goal is to find spins of the particles which satisfy these constraints as much as is possible. The couplings are fixed and the spins are variable. A configuration of spins has an energy that is known as the Hamiltonian. The more constraints are satisfied, the lower the energy. Since one cannot satisfy all of these constraints simultaneously, there is frustration. The goal is to reduce the frustration.

The energy landscapes of spin glasses and Ising models have been studied extensively due to their intriguing properties. Like in the optimization landscapes of AI models, they have many critical points, most of them being saddle points.

The AI analog of frustration is the loss function. Given a minibatch, SGD needs to update parameters so that the loss is reduced. The weights of the AI model are shared by many computation paths starting from different inputs and this introduces mutual constraints. Therefore, an intuitive analogy exists between spin glasses and AI models.

An example of a spin glass is the Ising model,
\[ H = - \sum_{i,j} J_{ij} \sigma_i \sigma_j , \]
where $H$ is the Hamiltonian, the couplings $J_{ij}$ are Gaussian random variables, and $\sigma_i = \pm 1$ are spins (degrees of freedom). If $J_{ij}>0$, then the spins need to align. If $J_{ij}<0$, then the spins need to be opposite. If $J_{ij}=0$, then there is no interaction. The couplings are between neighboring particles and there is a 2-D spatial structure.
This model can be generalized to a $p$-spin spherical model,
\[ H = - \sum_{i_1 > \ldots > i_p = 1}^N J_{i_1\ldots i_p} \sigma_{i_1} \ldots \sigma_{i_p} \]
where
\[ \sum_{i=1}^N \sigma_i^2 = N\]
is the spherical constraint on real-valued spins. Now the Hamiltonian is the result of $p$ particle interactions, all of which interact with all others, and there is no spatial structure in terms of neighbors. Note that $H$ is a polynomial in spins, just as an AI model is a polynomial in its learnable parameters. The couplings for spin glasses take the place of the input variables in the AI model.

For spherical spin glasses, it is possible to perform a mathematical analysis and discover an interesting layered structure of eigenvalues. Recall that index is the number of negative eigenvalues of the Hessian matrix. The statistics of critical points of index $k$ are related to the statistics of the $(k+1)$-th smallest eigenvalue. What can be shown is as follows.
\begin{itemize}
    \item At the bottom of the optimization landscape, there are only index 0 (local minima) critical points in a narrow band.
    \item In a band just above the bottom, there are local minima and saddle points with index 1.
    \item In a band above the second one, there are critical points with index 0, 1, 2.
    \item This layered structure continues with higher-index critical points appearing as the loss increases away from the bottom.
\end{itemize}
See~\cite{spin_glasses_auffinger_2013} for details.
Under certain assumptions, one can apply these results to AI, see~\cite{Choromanska_loss_surface_2015}. The learnable parameters take the role of the $p$-spins, the loss is the Hamiltonian, and the couplings are given by the input minibatch. This can be seen by noticing the similarity between the monomial terms:
\[J_{i_1\ldots i_p} \; \sigma_{i_1} \ldots \sigma_{i_p},\]
\[x_{i} \;w_{i_1} \ldots w_{i_d}.\]
The intuition is that when the loss is high, index is high, several eigenvalues are negative, and there are many directions in the saddle points to decrease the loss. Towards the bottom of the landscape the AI model is working very nicely, thus there are few degrees of freedom to improve it further.

\subsection{Computational Complexity Perspective}
There is an interesting perspective from computer science on the nature of high-dimensional optimization problems. NP-complete problems are intractable problems to solve because their worst-case time complexity increases exponentially with the problem size.

An example is the 3-SAT Boolean satisfiability problem, which asks whether a Boolean expression in 3-CNF (conjunctive normal form) with $M$ clauses and $N$ Boolean variables can be satisfied. An example for $M=3$ and $N=5$ is
\[ (x_1 \vee \overline{x_2} \vee x_3) \wedge (x_1 \vee x_3 \vee \overline{x_4}) \wedge (x_2 \vee x_4 \vee \overline{x_5}).\]
The decision problem is NP-complete. The optimization problem MAX-3SAT, which seeks to find the maximum number of clauses which can be satisfied, is NP-hard.
Parameterize the 3-SAT problem with a parameter,
\[\alpha = \frac{M}{N}.\]
Clauses represent `constraints' and Boolean variables represent `degrees of freedom', therefore the ratio $\alpha$ represents the difficulty of the problem. Using the analogy with spin glasses, clauses are like couplings and Boolean variables are like particles. Given an instance of 3-SAT one can reduce it to an instance of the graph problem of Maximal Independent Set. The vertices of the graph represent degrees of freedom, which will get colors, and edges represent constraints that no adjacent vertices get the same color. Therefore there is a direct analogy with spin glasses. Spin glasses are indeed closely connected to NP-complete problems\index{NP-complete problem}, see~\cite{spin_glasses_NP_complete}.

In fact, the connection is much deeper and this will further strengthen our belief that optimization in high-dimensional spaces is not intractable in practice.
One can randomly generate instances of 3-SAT for different values of $\alpha$ and compute the fraction of expressions which are unsatisfiable.
What can be shown is that as $\alpha \rightarrow 0$, most expressions are satisfiable. Intuitively, few constraints and many degrees of freedom make it easier to find a solution. As $\alpha$ increases, there is a phase transition when some expressions are satisfiable and some are not and then for larger values of $\alpha$ almost all expressions become unsatisfiable. This phase transition is similar to what has been observed in spin glasses, see~\cite{Kirkpatrick1994CriticalBI}.

Therefore, having many degrees of freedom leads to the solvability of hard problems. This is an important insight because in the case of AI, once we have fixed the size of the input, which represents `constraints', then by increasing the size of the network, which is the same as increasing the number of `degrees of freedom' of the AI model, one can satisfy these constraints.
Note that it is well known that fast approximation algorithms can provide near-optimal solutions to very large problem instances of NP-complete problems, which provides empirical justification of the presence of easily found, near-optimal solutions in high-dimensional landscapes.

\subsection{SGD and Critical Points}
\index{saddle point}
\index{critical point}
\index{stochastic gradient descent}
Since most of the critical points are saddle points in higher loss bands, we are unlikely to encounter a local minimum. How many saddle points\index{saddle point} will we encounter till we get stuck in a local minimum? We will work it out under the simplifying assumption that this can be formulated as a statistics question by modeling it as a geometric probability distribution. If the probability of a critical point being a local minimum is $p$, then the probability of encountering $t-1$ saddle points followed by a local minimum during SGD is
\[\Pr(X = t) = (1-p)^{t-1} p\]
with the expectation being
\[ \mathbb{E}(X) = \frac{1}{p} = \Omega(\exp(k n^2))\]
where $k \approx 0.275$. Recall that $\Omega$ refers to lower bound; the reciprocal of an upper bound gives a lower bound. This is an astronomically large number when $n$ is typically in the tens of millions. Therefore, it will be a long while before we encounter a local minimum. Note that $p$ is not constant but changes as we descend down the landscape. Since $p$ is smaller for earlier critical points, which belong to higher bands in the layered structure of the landscape, it is highly unlikely that we will get stuck in the higher bands.

Another question which arises is why SGD doesn't get stuck at a saddle point where the gradient is zero. The underlying reason is that SGD is a stochastic algorithm. It is unlikely that one will remain at a particular saddle point when new minibatches arrive. The landscape is fluid because the input changes and furthermore the underlying computation graph may change. Moreover, despite the gradient being zero, popular variations of SGD which keep a history of gradients can continue to move forward in the landscape. Saddle points may slow down SGD but in practice they don't seem to stop it altogether. See the paper~\cite{LeCun1998EfficientB} to see how the noisy nature of SGD is helpful.

\subsection{Confluence of Perspectives}
We built advanced intuition into tools which are available to us to answer the question --- why does an AI model get successfully trained? It is insightful to see a number of modern mathematical theories having rich interconnections.
\begin{enumerate}
    \item Theory of random polynomials. Since a ReLU neuron computes a polynomial, one can apply results from this branch of mathematics, which depend on results from the theory of random matrices.
    \item Theory of random matrices. Using Gaussian orthogonal ensembles, one can study statistics of eigenvalues of the Hessian matrix of the loss function, which also allows investigation of energy landscapes of spin glasses.
    \item Spin glasses and Ising models. These models formulate mutual constraints which free variables have to satisfy. The Hamiltonian of these physical systems is a random polynomial of spin variables.
    \item Computational complexity. Spin glasses are connected to NP-complete problems\index{NP-complete problem} in computer science, for which it is intractable to reach the global minimum. Yet fast approximation algorithms provide near-optimal solutions to very large problem instances.
\end{enumerate}
These different perspectives lead to the same insight that in high dimensions, the landscape has a rich mathematical structure which is amenable to algorithms such as SGD that have been fine-tuned to work very well in practice.

\printbibliography[heading=bibintoc, title={References}]
\end{document}